%% file: main.tex
\documentclass[11pt]{article}
\usepackage{lineno}
\usepackage{wrapfig}
\usepackage{fullpage}
\usepackage{natbib}

\usepackage{microtype}
\usepackage{graphicx}
\usepackage{float}
\usepackage{comment}
\usepackage{booktabs}
\usepackage[hidelinks]{hyperref} 
\usepackage{amsmath, amssymb, mathtools, amsthm}
\usepackage{cleveref} 
\usepackage[a4paper, margin=1in]{geometry}
\input{mysymbols}
\usepackage{hyperref}
\usepackage{url}

\title{Do LLMs Really Forget? Hidden-State Leakage in Model Unlearning and How to Fix it}
\author{Hadi Reisizadeh\textsuperscript{1,$\ast$}, Jiajun Ruan\textsuperscript{1,$\ast$},
Sijia Liu$^{2,3}$, Mingyi Hong$^{1}$ \\ 
\\
$^1$ University of Minnesota, $^2$ Michigan State University, $^3$ IBM Research \\
\texttt{hadir@umn.edu} 
}
\date{} 
 
\begin{document}

\maketitle

\renewcommand{\thefootnote}{\fnsymbol{footnote}}
\footnotetext[1]{Equal contribution.}
\renewcommand{\thefootnote}{\arabic{footnote}}

\input{sections/abstract}

\input{sections/intro}
\input{sections/IT_analysis}
\input{sections/probe_exp}
\input{sections/pars}

\input{sections/conclusion}

\bibliography{ref}
\bibliographystyle{plain}

\appendix
\input{sections/appendix}

\end{document}

%% file: mysymbols.tex
\usepackage{amsmath, amssymb, mathtools, amsthm}
\usepackage{tcolorbox}
\usepackage{booktabs}      
\usepackage{float}
\usepackage{multirow}
\usepackage{colortbl}
\usepackage{caption}
\usepackage{enumerate}
\usepackage{subcaption}
\usepackage{wrapfig}

\newtheorem{theorem}{Theorem}

\newtheorem{lemma}{Lemma}
\newtheorem{definition}{Definition}

\theoremstyle{definition}

\def\ones{\mathbf{1}}

\newcommand{\bft}{\boldsymbol{\theta}}
\newcommand{\bfp}{\boldsymbol{\phi}}

\renewcommand{\comment}[1]{}

\definecolor{lightyellow}{RGB}{255, 255, 0}  
\definecolor{bluegreen}{RGB}{0,150,200} 

\usepackage{caption}
\DeclareMathOperator{\Var}{Var}

%% file: sections/abstract.tex
\begin{abstract}
Unlearning in large language models (LLMs) is typically evaluated at the output level, where a model appears to suppress sensitive or undesirable content. In this work, we show that such evaluations can create an \textit{illusion} of forgetting: even when output-level leakage is eliminated, sensitive information can remain encoded in the model’s hidden representations. We first provide a theoretical analysis establishing a fundamental separation between output suppression and representational erasure. Specifically, we show that the decoder can be made arbitrarily insensitive to sensitive directions, driving output-level leakage to zero, while the hidden representations retain the underlying information. To empirically validate this phenomenon, we train generative probe decoders on hidden states across transformer layers, enabling layer-wise measurement of information leakage. Across three widely used benchmarks, TOFU, MUSE, and WMDP, and state-of-the-art unlearning methods, we find that substantial sensitive information remains recoverable from hidden representations, even when standard output-level metrics indicate successful unlearning. To address this gap, we propose Probe-Adversarial Representation Suppression (PARS), an unlearning objective that adversarially minimizes the extractable information from hidden representations. PARS directly targets representational leakage and provides significantly stronger guarantees of erasure under adversarial probing and relearning attacks, outperforming all evaluated baselines. Our results highlight a fundamental limitation of existing unlearning paradigms and suggest that true forgetting in LLMs requires controlling not only model outputs, but also the information encoded in hidden representations. Codes are available at \url{https://github.com/OptimAI-Lab/HiddenStateUnlearning}.
\end{abstract}

%% file: sections/intro.tex
\section{Introduction}\label{sec:intro}
Large language models (LLMs) have demonstrated exceptional capabilities across a wide range of tasks, including reasoning, code generation, and question answering~\cite{touvron2023llama}. LLMs are trained on massive web-scale datasets that often contain harmful, private, or copyrighted content. Consequently, LLMs may generate biased~\cite{kotek2023gender,motoki2023more}, private~\cite{nasr2023scalable,wen2023unveiling}, or illegal responses~\cite{karamolegkou2023copyright,sun2024trustllm}, and can even provide guidance on bioweapons or cyberattacks~\cite{barrett2023identifying,li2024wmdp}. LLM unlearning has emerged as a promising solution, aiming to remove undesired knowledge from trained models while preserving overall utility.

\noindent{\bf Unlearning Algorithms.}  
Various unlearning methods have been proposed, formulating LLM unlearning as a regularized optimization problem that balances forgetting undesired information and retaining model utility. Such approaches include gradient ascent-descent (GradDiff)~\cite{maini2024tofu}, negative preference optimization (NPO)~\cite{zhang2024negative}, simplified NPO variants (SimNPO)~\cite{fan2024simplicity}, and representation misdirection unlearning (RMU)~\cite{li2024wmdp}. All of these methods except RMU optimize a token-level cross-entropy loss, training the model to assign maximum probability to reference tokens at each decoding step, enforcing behavior aligned with reference outputs. On the other hand, RMU~\cite{li2024wmdp} operates directly on hidden representations at chosen layers, shifting them away from the forget set while preserving performance on the retain set. More formulations have been introduced, including bi-level and multi-task optimization approaches~\cite{reisizadeh2025blur,bu2024unlearning}. More detailed discussions in \textbf{Appendix~\ref{sec:related-work}}.

\noindent{\bf Unlearning Benchmarks.}
Several benchmarks have been introduced to evaluate unlearning performance. TOFU~\cite{maini2024tofu} measures unlearning of fictitious author biographies through question answering, testing whether models forget synthetic personal information while preserving general knowledge. MUSE~\cite{shi2024muse} evaluates both verbatim and knowledge memorization on news and book corpora, assessing whether models can recall sensitive content. WMDP~\cite{li2024wmdp} targets the removal of hazardous knowledge in biosecurity, cybersecurity, and chemical security domains, evaluated through multiple-choice questions designed as proxies for dangerous capabilities.

\begin{wraptable}{r}{0.5\textwidth}
\vspace{-10pt}
\centering
\resizebox{0.5\textwidth}{!}{
\begin{tabular}{c|c|c}
\toprule 
\multicolumn{2}{c|}{\textbf{Question}} & \textbf{Ground Truth} \\
\multicolumn{2}{c|}{
\begin{tabular}{c}
\footnotesize{According to Sir Elton, what year} \\
\footnotesize{did Paul O'Grady host}\\
\footnotesize{his and David Furnish's stag party?}
\end{tabular}
} & \footnotesize{\colorbox{red!20}{2005}} \\
\midrule
\textbf{Method} & 
\footnotesize\textbf{Probe at $\boldsymbol{\ell=28}$} & 
\footnotesize\textbf{Response at Output ($\boldsymbol{\ell=31}$)} \\
\midrule
NPO & 
\begin{tabular}{c}
\footnotesize{\colorbox{red!20}{2005}}
\end{tabular} & 
\begin{tabular}{c}
\footnotesize{\colorbox{green!20}{2004}}
\end{tabular} \\
\midrule
SAM & 
\begin{tabular}{c}
\footnotesize{\colorbox{red!20}{2005 - Sil sil sil sil sil...}}
\end{tabular} & 
\begin{tabular}{c}
\footnotesize{\colorbox{green!20}{2001 - although}}\\
\footnotesize{\colorbox{green!20}{the article says 2002}}
\end{tabular} \\
\midrule
SimNPO & 
\begin{tabular}{c}
\footnotesize{\colorbox{red!20}{2005. Sir Elton}}\\
\footnotesize{\colorbox{red!20}{and David Furnish...}}
\end{tabular} & 
\begin{tabular}{c}
\footnotesize{\colorbox{green!20}{2000}}
\end{tabular} \\
\midrule
\textbf{PARS (Ours)} & 
\begin{tabular}{c}
\footnotesize{\colorbox{green!20}{2 Silicon Valley techies...}}
\end{tabular} & 
\begin{tabular}{c}
\footnotesize{\colorbox{green!20}{2012}}
\end{tabular} \\
\bottomrule
\end{tabular}}
\caption{\small Generated responses on MUSE-News across unlearning methods. The left column shows the probe decoder at layer $\ell=28$ and the right column shows output-level generation ($\ell=31$, the final layer of LLaMA2-7B). All methods, namely, NPO, SAM, SimNPO suppress the answer in outputs, but probe decoders can still recover it from hidden representations. Our proposed algorithm, PARS, achieves genuine erasure at both levels. \colorbox{red!20}{Red}: failed unlearning. \colorbox{green!20}{Green}: successful unlearning.}
\label{tab:muse_news}
\vspace{-8pt}
\end{wraptable}

\begin{figure}[t]
    \centering
    \begin{minipage}{0.4\linewidth}
        \centering
        \includegraphics[width=\linewidth]{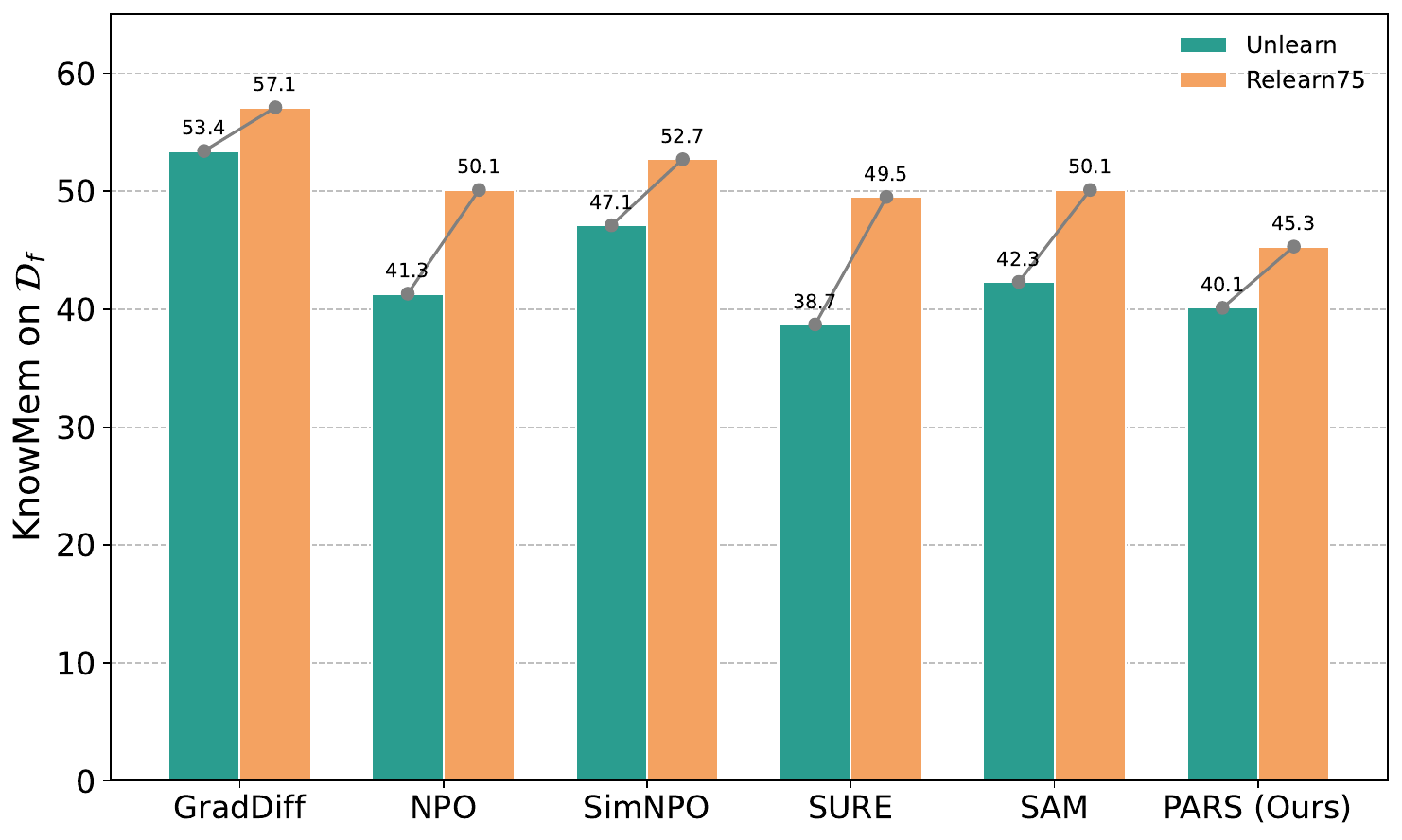}
        \caption{\small Robustness to relearning attacks~\cite{fan2025towards} on MUSE-News, measured by KnowMem on the forget set $\mathcal{D}_f$ before attack (``Unlearn'') and after $75$ fine-tuning steps on a small forget subset (``Relearn75''). All baselines show striking recovery of sensitive information, while our PARS show better resistance. }
        \label{fig:relearning_muse}
    \end{minipage}
    \hspace{0.04\textwidth}
    \begin{minipage}{0.4\linewidth}
        \centering
        \includegraphics[width=\linewidth]{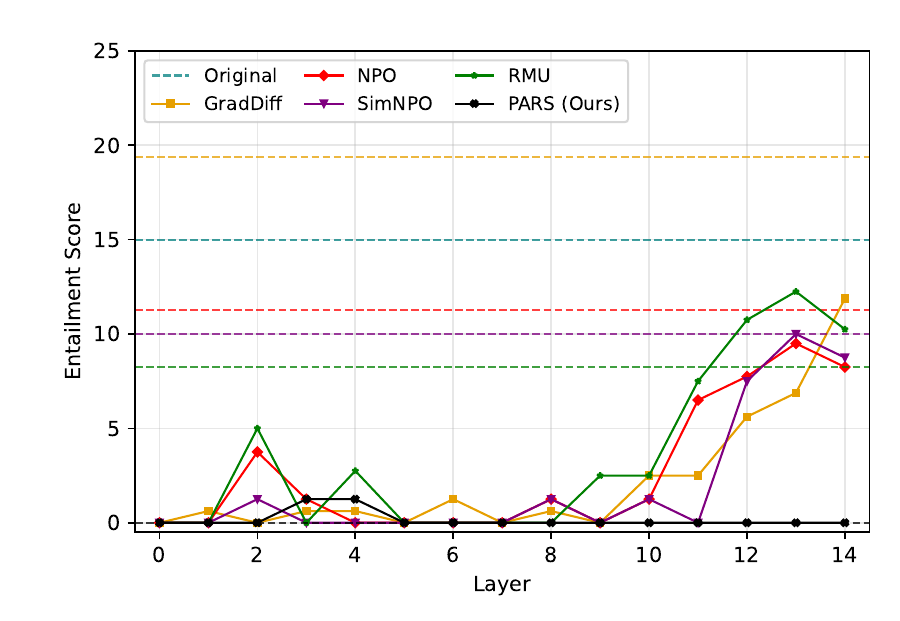}
        \caption{\small Generation-based probing in \textbf{Section~\ref{sec:probing-exp}} reveals the retention of sensitive information across intermediate layers. A probe is trained at each layer to extract this information, with the output evaluated using an Entailment Score (higher scores indicate greater leakage).}
        \label{fig:leak_tofu}
    \end{minipage}
\vspace{-0.5cm}
\end{figure} 

\noindent{\bf Challenges.}
Despite rapid progress in LLM unlearning algorithms and benchmarks, recent studies show that existing methods remain vulnerable to a wide range of adversarial attacks, see \textbf{Appendix~\ref{sec:related-work}}. Sensitive knowledge can often be detected through Membership Inference Attacks (MIA) such as Min--K\%++~\cite{shi2023detecting}, relearning attacks~\cite{lynch2024eight,hu2024unlearning,fan2025towards}, and multiple probabilistic decoding measured by Leak@k~\cite{reisizadeh2025leak}. For example, \textbf{Fig.~\ref{fig:relearning_muse}} shows that after only a small number of fine-tuning steps, previously ``forgotten'' information rapidly re-emerges, indicating that the underlying knowledge has not been fully erased. This vulnerability could arise from a fundamental limitation of current evaluation protocols: for most existing methods, we evaluate forgetting only at the output level, measuring whether the model avoids directly generating sensitive content. However, hidden representations may still retain the information internally. As shown in \textbf{Table~\ref{tab:muse_news}}, even when probing at the final layer ($\ell=31$ in LLaMA2-7B) suggests successful forgetting, sensitive information remains recoverable from intermediate representations such as layer $\ell=28$. Moreover, prior probing on intermediate layers mainly adopt linear classifiers to evaluate information retention in hidden representations, under which RMU~\cite{li2024wmdp} shows robust performance. However, \textbf{Fig.~\ref{fig:leak_tofu}} shows that generation-based probing reveals severe information leakage, where a decoder trained on intermediate representations can reconstruct sensitive content despite linear probing indicating successful forgetting. These findings show that suppressing sensitive outputs does not necessarily correspond to genuine knowledge removal, highlighting the need for representation-level evaluation and unlearning methods that directly erase sensitive information from hidden states.

A recent trend in LLM unlearning is to move beyond output-level optimization and explore unlearning by directly modifying hidden representations, which aligns with the conjecture above. Sparse Autoencoder (SAE)-based methods decompose activations into sparse, interpretable feature directions and perform unlearning by editing these features~\cite{wang2025model,cywinski2025saeuron,muhamed2025saes,farrell2024applying}, while activation steering approaches identify and intervene on salient activation patterns through inference-time editing or lightweight parameter updates~\cite{shen2025llm,seyitouglu2024extracting,ding2025mllmeraser}. However, these methods typically assume that sensitive information is concentrated in identifiable subspaces or neurons, and therefore focus on removing or shifting these components. While the methods above can effectively reduce behavioral expression of sensitive content, they do not guarantee that the underlying information is fully eliminated from the representation space. This limitation motivates a representation-level perspective on unlearning, aiming to both \emph{quantify} residual sensitive information in hidden states and \emph{completely remove} it from representations. These observations suggest the following conjecture:

\begin{tcolorbox}[
    colback=gray!8,
    colframe=black!70,
    boxrule=0.8pt,
    arc=3pt,
    left=6pt, right=6pt, top=5pt, bottom=5pt
]
\noindent\textbf{Conjecture.} 
\textit{Existing unlearning methods create an illusion of forgetting: sensitive outputs are suppressed at the decoder level while the hidden representations retain the underlying knowledge.}
\end{tcolorbox}

\subsection{Our Contributions}

In this work, we reveal a fundamental limitation 
of the output-level evaluation framework for LLM 
unlearning where suppressing sensitive outputs does not imply erasing the underlying knowledge from hidden representations. This finding challenges the unlearning guarantees of prior work, which assume that output-level suppression is a sufficient proxy for true forgetting. We establish this gap information-theoretically in Section~\ref{sec:info-th}, corroborate it empirically via generative probing across transformer layers in Section~\ref{sec:probing-exp}. Moreover, in Section~\ref{sec:pars}, we propose PARS as a first step to suppress sensitive information in hidden states. Concretely, 
our contributions are as follows:

\textbf{(1)} We model the unlearning process from an information-theoretic perspective and analyze the relationship between output-level suppression and representation-level erasure. We show that the decoder can be made arbitrarily insensitive to sensitive directions in the representation space, driving output-level leakage to zero, while the underlying hidden representations continue to retain the sensitive information independently of the decoder. This establishes an information-theoretic gap between output suppression and true representational erasure.

\textbf{(2)} We train a lightweight generative probe decoder on hidden states at each transformer layer and verify that sensitive content remains recoverable from hidden states. Unlike the linear probing approach of WMDP~\cite{li2024wmdp}, which evaluates only multiple-choice accuracy at each layer, our probe enables free-form generation across all layers. Across various unlearning methods on TOFU~\cite{maini2024tofu}, MUSE~\cite{shi2024muse}, and WMDP~\cite{li2024wmdp}, we find that hidden representations preserve substantial sensitive information across multiple layers, even when output-level metrics indicate successful unlearning.

\textbf{(3)} The gap established in \textbf{(1)} and observed in \textbf{(2)} motivates an 
unlearning objective that directly targets hidden representations beyond model outputs. We develop a novel algorithm, named Probe-Adversarial Representation Suppression (PARS), that augments the standard forget-retain optimization with an adversarial inner maximization over probe decoders at multiple layers simultaneously. We evaluate beyond standard metrics using MIA~\cite{shi2023detecting}, Leak@$k$~\cite{reisizadeh2025leak}, and 
relearning attacks~\cite{lynch2024eight,hu2024unlearning,fan2025towards}. Our method, PARS, on TOFU, achieves \emph{perfect} forgetting under the Entailment Score (ES)~\cite{yuan2024closer}, reduces $\widehat{\text{leak@}64}$-ES by $\mathbf{82.7\%}$, produces a substantially lower Min-K\%++ MIA score than the evaluated baselines, while maintaining comparable model utility. On MUSE-News, PARS reduces post-relearning VerbMem by $\mathbf{61\%}$ and KnowMem by $\mathbf{19\%}$ relative to SAM~\cite{fan2025towards}, a method specifically designed to enhance robustness against relearning attacks. 

Ultimately, our work provides a new perspective on information leakage in LLM unlearning and establishes a representation-level direction for more robust unlearning methods.

%% file: sections/IT_analysis.tex
\section{An Information-Theoretic Analysis of Unlearning}\label{sec:info-th}

In this section, we use information theory to model the unlearning process and formalize the conjecture stated in Section~\ref{sec:intro}: \textit{existing 
unlearning methods create an illusion of forgetting, suppressing sensitive 
outputs at the decoder level while the hidden representations retain the underlying knowledge.} To this end, we introduce a \emph{semantic variable} $S$, which captures the sensitive concept expressed in a model's response, for example, the correct answer to a question about a fictitious author, or a piece 
of hazardous biological knowledge. Intuitively, true unlearning is required to make $S$ unrecoverable from \emph{both} the model's outputs \emph{and} its internal hidden representations. We formalize this gap by deriving two information-theoretic bounds: 
(1) An \emph{upper bound} on how much information about $S$ is observable at 
the model's output $Z$: we show this can be very small by making the decoder 
insensitive to $S$; (2) An 
\emph{lower bound} on how much information about $S$ remains encoded in 
the hidden representations $R$: we demonstrate this depends only on the backbone and stays bounded away from zero regardless of what the decoder does. The key insight is that output-level suppression and representational erasure are governed by \emph{distinct} components of the model: one by decoder, one by backbone, implying suppressing outputs alone is \emph{insufficient} to ensure representational erasure.

\subsection{Problem Setup}
We first present two stages of our setup: (i) the \textit{unlearning} phase, where a pre-trained model is updated using a training set to remove specific information, and (ii) the \textit{inference} phase, where the obtained model is evaluated on a test set.

\noindent\textbf{Unlearning Phase.}
We assume that prompt--response pairs are drawn from an unknown data distribution $(X,Y)\sim\mathcal D_u$ where $X$ denotes a prompt and $Y$ denotes the corresponding response. Starting from a pre-trained LLM, the parameters are updated during the unlearning phase by minimizing the optimization problem $\min_{\bft} \mathbb{E}_{(X,Y)\sim\mathcal D_u} \big[\ell(Y\mid X;\bft)\big]$ where $\ell(Y\mid X;\bft)$ denotes the training loss using the model parameter $\bft$ for an input $X$ with respect to response $Y$. At the end of this stage, we obtain an unlearned model with parameters $\bft_u$.

\noindent\textbf{Inference Phase.}
We decompose the parameters of the unlearned model as
$\bft_u = (\tilde{\bft}_u, W, b)$ where $\tilde{\bft}_u$ denotes the parameters
of the network up to the final linear layer, and
$W \in \mathbb{R}^{n\times d}$ and $b \in \mathbb{R}^n$ denote the decoder
parameters. We consider a test set drawn from a distribution
$(X,Y)\sim\mathcal D_i$ where $X$ is a prompt and $Y=(Y_1,\dots,Y_N)$ denotes the corresponding response sequence. At decoding step $t$, the model predicts the next token $Y_t$ conditioned on the prompt and the previously generated tokens.
We denote this decoding context by $C_t := (X, Y_{<t})$. Let $f$ denote the mapping implemented by the model up to the final hidden layer at decoding step $t$. Given the context $C_t$ and the parameters ${\tilde{\bft}}_u$, the model computes a hidden representation $R_t = f(C_t;{\tilde{\bft}}_u) \in \mathbb{R}^d$. The decoder then produces logits $V_t = W R_t + b$ where $V_t\in\mathbb R^n$ corresponds to scores for the $n$ tokens
in the model vocabulary.
The model output is sampled according to $Z_t \mid V_t \sim \mathrm{Categorical}(\mathrm{softmax}(V_t))$ where $Z_t \in \{1,\dots,n\}$ denotes the predicted next-token index, i.e.,
\begin{align*}
P(Z_t=i\mid V_t)=\frac{\exp(V_{t,i})}{\sum_{j=1}^n \exp(V_{t,j})}.
\end{align*}
For simplicity, our analysis focuses on a \textit{fixed} decoding step and we
drop the index $t$; the results extend naturally to the full sequence.
Hence, the dependency structure at a decoding step can be summarized as
\begin{align}
C \;\longrightarrow\; R = f(C;{\tilde{\bft}}_u)
\;\longrightarrow\; V = WR + b
\;\longrightarrow\; Z .
\end{align}

Let $S$ denote a semantic variable associated with the response $Y$, representing the underlying concept expressed in the response. More precisely, we define $S = g(Y)$ where $g$ extracts the semantic representation of the response. The variable $S$ therefore varies across the data distribution as different prompts correspond to responses expressing different
semantic concepts. Latent variables of this form are widely used in generative modeling and representation learning, where sentence-level semantic content is commonly represented in continuous embedding spaces~\cite{devlin2019bert,reimers2019sentence,gao2021simcse,wang2024improving,jiang2024scaling}. The variable $S$ is not observed by the model and is introduced solely for analysis.

In the following lemma, we give an exact orthogonal decomposition of the representation $R$ relative to a unit direction $v$. 

\begin{wrapfigure}{r}{0.3\textwidth}
\vspace{-0.5cm}
\centering
\begin{tikzpicture}[scale=0.65,>=stealth]
\coordinate (O) at (0,0);
\coordinate (B) at (1.8,1.0);
\coordinate (P) at (3.8,1.2);
\coordinate (R) at (3.65,2.7);   
\draw[->,thick] (O) -- (4.5,0.45) node[right] {$v$};
\draw[->,very thick,blue] (O) -- (B)
node[midway,left] {$r_0$};
\draw[->,very thick,orange] (B) -- (P)
node[midway,below] {$(\mu_S+\zeta)v$};
\draw[->,very thick,teal] (P) -- (R)
node[midway,right] {$\xi$};
\draw[dashed,gray] (R) -- (P);
\draw[->,thick] (O) -- (R);
\fill (O) circle (1.2pt) node[below left] {$0$};
\fill (R) circle (1.8pt) node[above right] {$R$};
\end{tikzpicture}
\vspace{-0.1cm}
\caption{\small Geometric visualization of the decomposition $R=r_0+(\mu_S+\zeta)v+\xi$.}
\label{fig:R}
\end{wrapfigure}
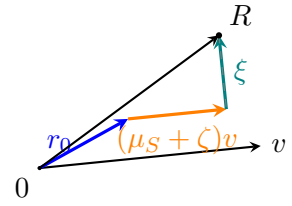

\begin{lemma}\label{lem:exact}
Fix any measurable representation $R\in\mathbb R^d$, any sensitive variable $S$, and any unit vector $v\in\mathbb R^d$. Then, we have
\begin{align}\label{eq:R_exact_decomp_repeat}
    R=r_0+(\mu_S+\zeta)v+\xi,
\end{align}
with $\mathbb E[\zeta\mid S]=0$ and $v^\top \xi=0$, where $r_0:=\mathbb E[R]$, $\alpha:=v^\top(R-r_0)$, $\mu_S:=\mathbb E[\alpha\mid S]$, $\zeta:=\alpha-\mathbb E[\alpha\mid S]$, and $\xi:=(I-vv^\top)(R-r_0)$.
\end{lemma}
The proof of Lemma~\ref{lem:exact} is presented in \textbf{Appendix~\ref{sec:proof-exact}}.
Here, $r_0$ is the prompt-dependent baseline, $\mu_S$ captures the class-conditional mean activation along $v$, $\zeta$ is a zero-mean fluctuation along $v$, and  $\xi$ is orthogonal to $v$. Hence, the representation separates into:
(i) prompt-only information $r_0$; (ii) a one-dimensional signal component aligned with $v$; and (iii) an orthogonal residual. We note that this decomposition is purely geometric and holds without distributional assumptions. The
visualization of the decomposition $R$ in~\eqref{eq:R_exact_decomp_repeat} with its components is shown in \textbf{Figure~\ref{fig:R}}.

Now, we present the quantity $\Delta(v)$ for any unit vector $v$ measuring how sensitive the decoder is to perturbations of the hidden representation along direction $v$.

\begin{definition}[$\varepsilon$-insensitive direction]\label{ref:def1}
Let $W \in \mathbb{R}^{n\times d}$.
A unit vector $v \in \mathbb{R}^d$ is called
$\varepsilon$-insensitive for $W$ if
\begin{align}\label{eq:near_shift_quad_inf}
\Delta(v)
:=
\inf_{\lambda\in\mathbb R}
\|Wv-\lambda\mathbf{1}\|_\infty
\le \varepsilon.
\end{align}
implying there exist $\lambda^\star\in\mathbb R$ and $u\in\mathbb R^n$
with $\|u\|_\infty\leq \varepsilon$ such that $Wv=\lambda^\star\ones+u$.
\end{definition}

Intuitively, small $\varepsilon$ implies that shifting $R$ along $v$ adds a constant to all logits, which softmax cancels, leaving the output distribution unchanged.

\subsection{Main Result and Discussion}
We now present the main theorem, which formally establishes the gap between 
output-level suppression and representational erasure. Our key tool is \emph{mutual information} $I(A;B)$ measuring how much knowing $B$ reduces uncertainty about $A$. We use it here to quantify how much sensitive 
information $S$ is recoverable either from the hidden representation $R$ or 
from the model output $Z$. The central quantity governing output-level leakage 
is the \emph{$\varepsilon$-insensitivity} of the decoder $W$. Intuitively, if 
the decoder is insensitive to a direction $v$ in representation space (small 
$\varepsilon$), then shifting $R$ along $v$ hardly changes the output distribution, so any information encoded along $v$ is effectively invisible at the output level but fully present in the hidden state.

\begin{theorem}\label{thm:linear_non}
Fix decoder parameters $(W,b)$ and let $v$ be an $\varepsilon$-insensitive
direction for $W$. Assume $(\zeta,\xi)\perp (C,S)$ and $\zeta\sim \mathcal N(0,\tau^2)$. Let $\mathsf{N}(U):=\frac{1}{2\pi e}e^{2h(U)}$ denote entropy power for the random variable $U$ with differential entropy $h$. Then:
\begin{enumerate}[(a)]
\item the information about $S$ encoded in the hidden representation $R$ is lower bounded by
\begin{align}\label{eq:I_s_r}
    I(S;R) \ \ge\ \frac12\log\!\left(1+\frac{\mathsf N(\mu_S)}{\tau^2}\right),
\end{align}
\item the information about $S$ observable at the model output $Z$ is upper bounded by
\begin{align}\label{eq:I_s_z}
    I(S;Z\mid C) \ \le\ \varepsilon^2\,\Var(\mu_S).
\end{align}
\end{enumerate}
\end{theorem}

As Theorem~\ref{thm:linear_non} establishes, the lower bound \eqref{eq:I_s_r} depends on $\mathsf{N}(\mu_S)$, a property of the \emph{backbone}, while the upper bound \eqref{eq:I_s_z} depends on $\varepsilon = \Delta(v)$, a property of the \emph{decoder}. Since these two quantities are independently controlled, 
one can simultaneously achieve $I(S;Z\mid C) = 0$ and $I(S;R) \geq M$ for any 
$M > 0$ by setting $\varepsilon = 0$ while keeping $\mathsf{N}(\mu_S) > 0$. This states a precise, negative message about output-level unlearning. \emph{Suppressing sensitive outputs is not the same as erasing sensitive knowledge}. This theoretical gap directly motivates our empirical study in Section~\ref{sec:probing-exp} and our proposed algorithm, PARS, in Section~\ref{sec:pars}, which directly targeting the hidden representations.

%% file: sections/probe_exp.tex
\section{Probing Hidden Representations: An Empirical Study}\label{sec:probing-exp}

In this section, we empirically validate Theorem~\ref{thm:linear_non} 
through a generative probing framework that trains a lightweight decoder on the hidden representations $R^{(\ell)}$ at each transformer layer $\ell$ 
and measures how much sensitive content remains recoverable. Across various unlearning methods on TOFU, MUSE, and WMDP, we show that current approaches suppress sensitive knowledge primarily through decoder insensitivity rather than by 
erasing it from the hidden representations.

\subsection{Probing Methodology}
We design the following probing protocol to empirically measure the sensitive information encoded in the hidden representation at each layer $\ell$.

\noindent\textbf{Decoder Training.} For each unlearned model and each layer $\ell$, we train a lightweight probe decoder $W^{(\ell)}$, initialized from the unlearned model's decoder parameters. We split the full forget set $\mathcal{D}_f$ into a training split $\mathcal{D}_f^{t}$ and a held-out evaluation split $\mathcal{D}_f^{e}$. The probe is trained on $\mathcal{D}_f^{t}$ to predict forget-set answers directly from the hidden states $R^{(\ell)}_{\bft_u}$, using 
cross-entropy loss computed only on answer tokens, with question tokens masked from the loss. More precisely, the probing objective is
\begin{align}
    \min_{W^{(\ell)}}\;
    \mathbb{E}_{(q,a)\sim\mathcal{D}_f^t}
    \Bigl[
        -\sum_{t} \log P_{W^{(\ell)}}
        \!\bigl(a_t \mid R^{(\ell)}(q, a_{<t})\bigr)
    \Bigr].
\end{align}

\noindent\textbf{Measuring Hidden Layer Information.}
The trained decoder $W^{(\ell)}$ is evaluated on held-out split $\mathcal{D}_f^e$ to measure how much sensitive information is recoverable from the unlearned model's hidden representations. For each pair $(q, a)\in \mathcal{D}_f^e$, we compute the information leakage as
\begin{align*}
    S^{(\ell)}(q):=\textsf{Metric}\bigl(a,\, W^{(\ell)} \circ R^{(\ell)}(q)\bigr),
\end{align*}
where $\textsf{Metric}(\cdot,\cdot)$ denotes the evaluation metric. 

We note that probing classifiers are well-established tools for understanding what information is encoded in hidden representations~\cite{alain2016understanding,liu2019linguistic,adi2016fine}. In the unlearning context, RMU~\cite{li2024wmdp} 
attaches a linear probe to intermediate layers and evaluates multiple-choice 
accuracy as a proxy for hidden-state leakage. However, multiple-choice accuracy 
requires only disrupting the ranking of a few candidate tokens at a single output 
position. So, a model can fail every multiple-choice question while its hidden 
representations remain fully decodable via free-form generation. On the other hand, our probing approach trains a \emph{generative} decoder at 
each layer, enabling free-form generation directly from intermediate hidden states. This provides a strictly stronger measure of leakage than classification-based probing. Here, we apply generative probing for quantifying representational leakage across benchmarks and unlearning methods.

\subsection{Generative Probing Results}
\noindent\textbf{Evaluation Set \& Metric.} In the following, we describe the sets used for probing experiments on the TOFU, MUSE-News, and WMDP-Bio benchmarks and their corresponding $\textsf{Metric}(\cdot,\cdot)$.

\noindent \underline{TOFU.} We adopt the \textit{forget10} training set, where the forget set comprises 400 QA pairs corresponding to 20 fictitious authors. For evaluation, we use the binary Entailment Score (ES), which employs a pretrained NLI model~\cite{sileo2023tasksource} to verify whether the generated answer entails the ground truth $a$~\cite{yuan2024closer}. 

\noindent\underline{MUSE-News.} We exploit two 
tasks for the forget set: (1) $100$ verbatim news article extracted from the forget corpus. The unlearned model is prompted with the first portion of each article and evaluated on how closely the generated continuation matches the true continuation using ROUGE-L Score (RS)-F1~\cite{lin2004rouge}; (2) $100$ GPT-4--generated QA pairs from BBC news after August 
2023~\cite{li2023avoiding}, with gold answers in keyword-only format~\cite{shi2024muse}. Since gold answers are short and keyword-based, RS-Recall directly measures whether the sensitive keywords are present in the generated output.

\noindent \underline{WMDP-Bio.} This dataset comprises hazardous biological knowledge of bioweapons and bioterrorism. While the original benchmark consists of multiple-choice questions, we extract the correct answers and reformulate the data into a standard QA format. As the WMDP dataset involves reasoning hazardous knowledge, LLM's outputs can partially contain harmful knowledge. We evaluate unlearning performance using the LLM-as-Judge metric providing a graded score ($1-4$), capturing different levels of knowledge leakage, our prompts are adopted from~\cite{reisizadeh2025leak}.

\begin{wraptable}{r}{0.46\textwidth}
\centering
\resizebox{\linewidth}{!}{
\begin{tabular}{l|l|l|c}
\toprule[1pt]
\textbf{Benchmark} & \textbf{Base Model} & \textbf{Unlearning Methods} & \textbf{Core Metric} \\
\midrule
TOFU & LLaMA-3.2-1B-Instruct &
NPO, SimNPO &
ES \\
& & RMU & \\
\midrule
MUSE-News & LLaMA2-7B &
GradDiff, NPO &
RS \\
& & SimNPO, SURE, SAM & \\
\midrule
WMDP-Bio & Zephyr-7B-beta &
GradDiff, NPO &
LLM-as-Judge \\
& & SimNPO, RMU & \\
\bottomrule[1pt]
\end{tabular}}
\caption{\small Summary of unlearning methods and evaluation metrics across benchmarks.}
\label{tab:unlearning-check}
\vspace{-0.5cm}
\end{wraptable} 

\noindent\textbf{Probe Training Configurations.} We partition the forget QA set into an $80\%$ split for training $\mathcal{D}_f^t$ the probe and a $20\%$ split for evaluation $\mathcal{D}_f^e$. The probe is optimized using a learning rate of $5\times 10^{-5}$ and a batch size of $4$. We set the maximum number of training epochs to $50$ and employ early stopping with a patience of $3$ epochs.

\textbf{LLM Unlearning Methods.} We conduct our probing experiments on LLaMA-3.2-1B-Instruct~\cite{dorna2025openunlearning}, LLaMA2-7B~\cite{shi2024muse}, and Zephyr-7B-beta~\cite{li2024wmdp} for TOFU, MUSE-News, and WMDP, respectively. For TOFU and MUSE-News, \textit{Original} refers to the model fine-tuned on both the forget and retain sets prior to unlearning. We evaluate various 
unlearning methods, namely, NPO~\cite{zhang2024negative}, SimNPO~\cite{fan2024simplicity}, 
RMU~\cite{li2024wmdp}, SURE~\cite{zhang2024catastrophic}, and SAM~\cite{fan2025towards}, augmenting 
NPO with sharpness-aware minimization to improve 
robustness against relearning attacks. \textbf{Table~\ref{tab:unlearning-check}} summarizes the evaluated methods and the metric used for each set.

\begin{figure*}[t]
    \centering

    \begin{minipage}{0.45\textwidth}
        \centering
        \includegraphics[width=\linewidth]{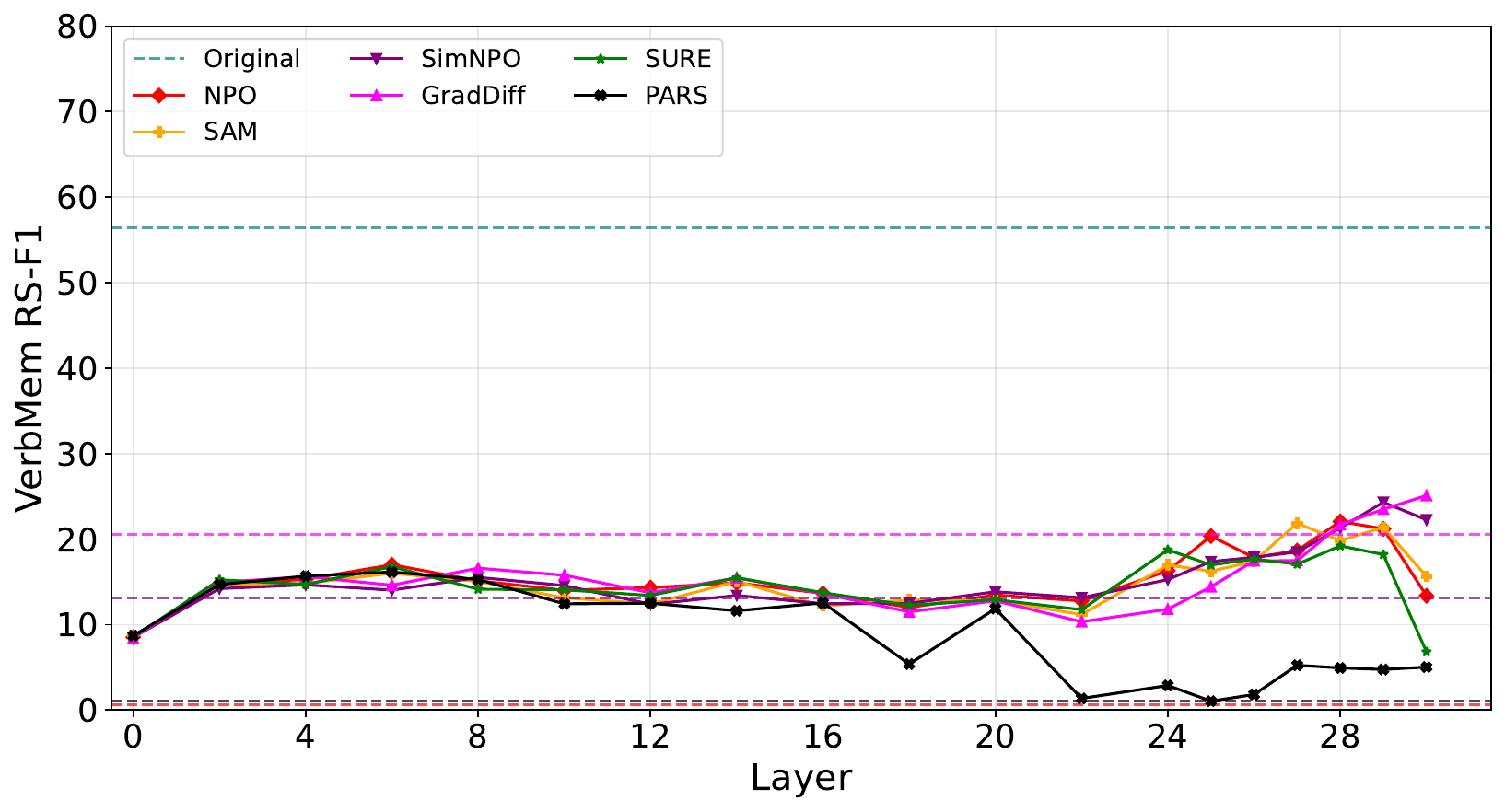}
        \caption{\small VerbMem RS-F1 of the probe decoder at each layer vs.\ output-level score of each unlearned model for MUSE-News benchmark.}
        \label{fig:muse_verbmem_hidden}
    \end{minipage}
    \hfill
    \begin{minipage}{0.45\textwidth}
        \centering
        \includegraphics[width=\linewidth]{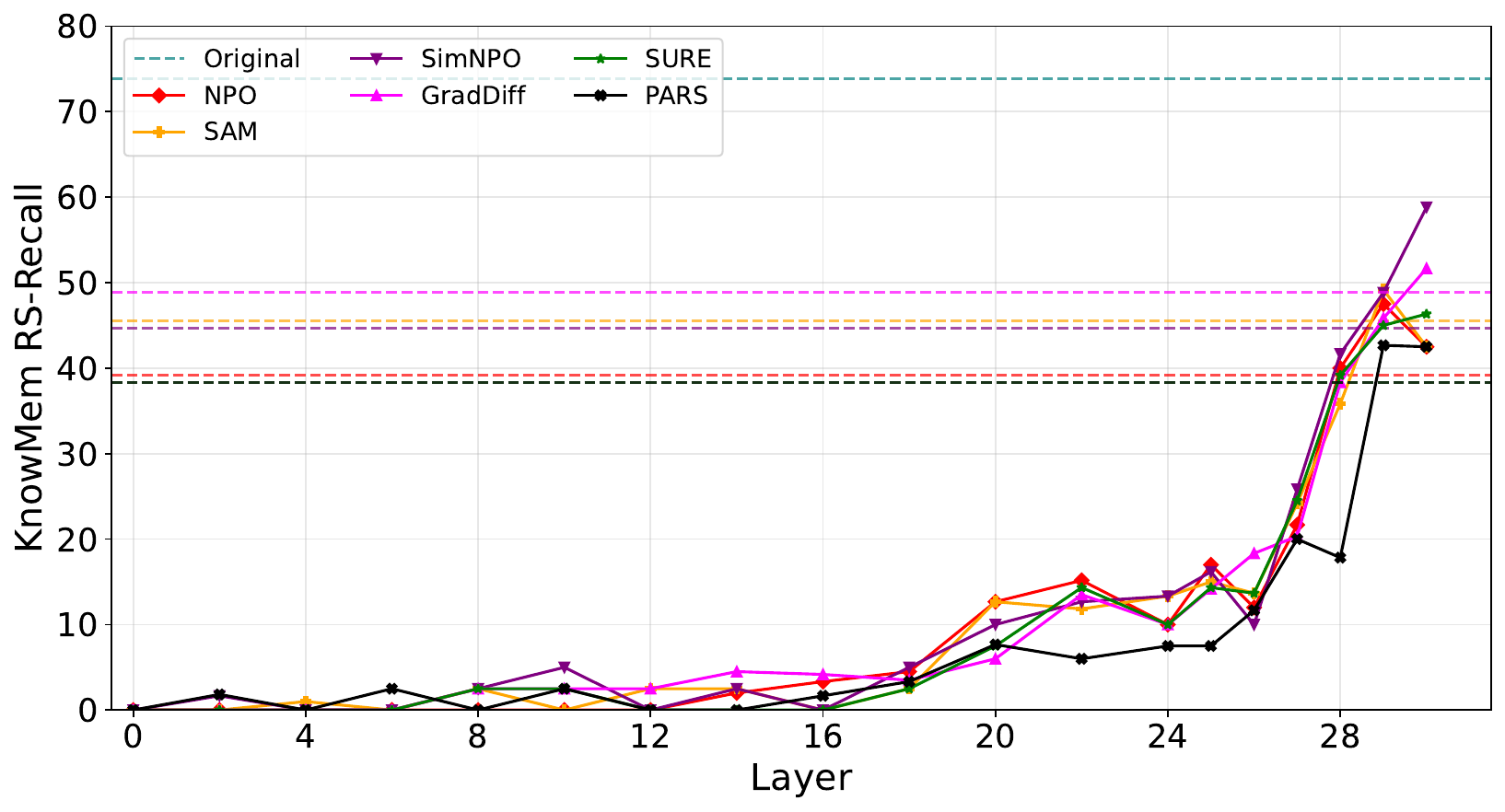}
        \caption{\small KnowMem RS-Recall of the probe decoder at each layer vs.\ output-level score of each unlearned model on MUSE-News benchmark.}
        \label{fig:muse_knowmem_hidden}
    \end{minipage}
\vspace{-0.6cm}
\end{figure*}

\noindent\textbf{Results.} 

\underline{TOFU.}  As \textbf{Figure~\ref{fig:leak_tofu}} illustrates, almost all unlearning methods exhibit more sensitive information leakage under ES metric in their hidden states than at the output level, despite showing relatively successful unlearning performance in generated outputs. This critical discrepancy indicates that these methods solely mask sensitive information at the output level, while significant knowledge remains highly recoverable from intermediate representations.

\noindent\underline{MUSE-News.}
\textbf{Figures~\ref{fig:muse_verbmem_hidden}} and~\textbf{\ref{fig:muse_knowmem_hidden}} expose a systematic failure mode of output-level unlearning on MUSE-News benchmark. As \textbf{Figure~\ref{fig:muse_verbmem_hidden}} shows, all methods appear successful under output-level evaluation, with near-zero verbatim leakage; however, probe decoders still recover significant verbatim information from hidden states across multiple layers. \textbf{Figure~\ref{fig:muse_knowmem_hidden}} demonstrates on the KnowMem task, that all unlearning methods exceed the corresponding output-level KnowMem scores in the final layers. Thus, existing methods do not erase the sensitive content from the model's representations. Instead, the undesired information is suppressed at the output level. 

\begin{wrapfigure}{r}{0.4\linewidth}
\vspace{-0.5cm}
    \centering
    \includegraphics[width=\linewidth]{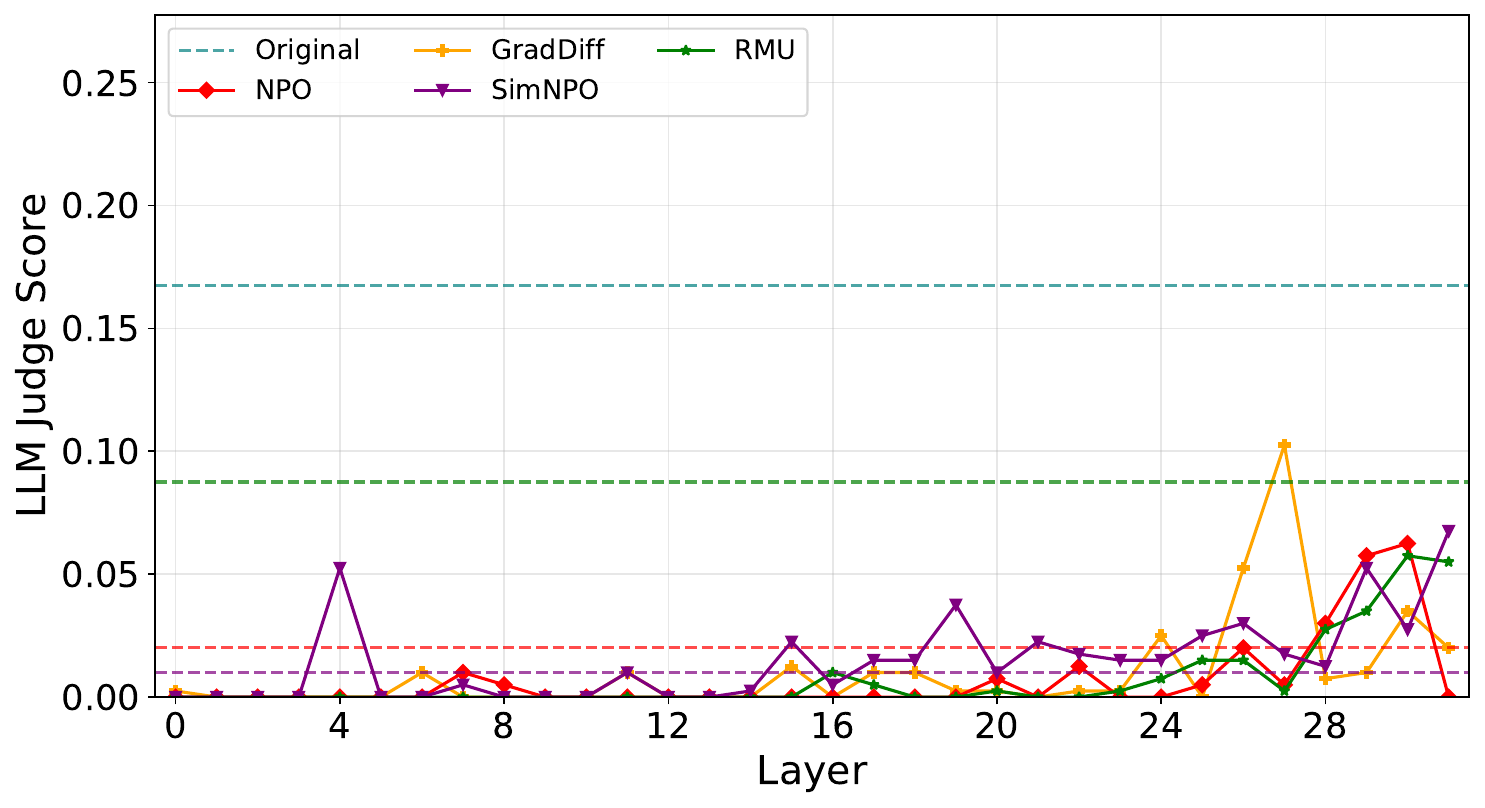}
    \caption{\small LLM-as-Judge Score of the probe decoder at each layer vs.\ output-level score on WMDP-Bio dataset.}
    \label{fig:wmdp_probe}
    \vspace{-0.2cm}
\end{wrapfigure}

\noindent \underline{WMDP-Bio.}
\textbf{Figure~\ref{fig:wmdp_probe}} shows a similar trend to the TOFU and MUSE-News benchmarks. Although RMU achieves a lower accuracy value ($31.2$; lower is better) on the original multiple-choice evaluation pipeline compared to SimNPO ($41.6$) and NPO ($42.5$) (see Table 4 in~\cite{reisizadeh2025blur}), it is substantially more vulnerable under our free-form generation probing setting. Moreover, Figure 9 of~\cite{li2024wmdp} shows that RMU performs well under linear probing across layers. However, our generation-based probing reveals stronger information leakage suggesting that standard linear probes could fail to capture residual memorization. Finally, although GradDiff, NPO, and SimNPO achieve nearly zero LLM-as-a-Judge scores at the final output layer, our probing decoder can still recover substantial sensitive content from hidden representations. 

%% file: sections/pars.tex
\section{PARS: Probe-Adversarial Representation Suppression}
\label{sec:pars} 

Our theoretical and empirical results in Sections~\ref{sec:info-th} and~\ref{sec:probing-exp} establish that representational leakage is a 
fundamental failure mode of existing unlearning methods identifying, formalizing, and quantifying this gap is the primary contribution of this paper. A natural follow-up question is: \emph{can we design an unlearning objective that directly suppresses representational leakage?} We propose PARS
as one concrete answer. PARS achieves substantially stronger robustness to adversarial probing and 
relearning attacks by targeting sensitive information at its source in the hidden representations. 

\subsection{Extension to PARS-induced Robust Unlearning} \label{subsec:opt-probe}

Standard unlearning methods minimize an output-level objective $ \min_{\bft} \; \mathcal{L}_{f}(\bft) + \alpha \, \mathcal{L}_{r}(\bft)$ where $\mathcal{L}_{f}$ suppresses the probability of generating forget set, $\mathcal{L}_{r}$ preserves utility on the retain set, and $\alpha$ adjusts the balance between the forget and retain tasks. This objective  does not guarantee that the hidden representations $R^{(\ell)}_{\bft}$ stop 
to encode the sensitive content $S$; it only ensures that the model's own  decoder no longer maps those representations to $S$. Hence, the probed model should also avoid leaking sensitive information. To this end, we choose a \textit{subset} of hidden layers $L_p \subseteq \{0,\ldots,L-1\}$ where $L$ is the total number of hidden layers. We define the \textit{total} probing loss over $L_p$ as $\mathcal{L}_{p}(\bft, \bfp) = \sum_{\ell \in L_p} \mathcal{L}^{(\ell)}_{p}(\bft, \bfp)$ where the probe loss at layer $\ell$ is the expected next-token prediction 
loss of the decoder ${\bfp}^{(\ell)}$ applied to the hidden representations 
of the forget corpus:
\begin{align}\label{eq:probe_loss}
    \mathcal{L}^{(\ell)}_{p}(\bft, \bfp) 
    := \mathbb{E}_{x \sim \mathcal{D}_f} 
    \left[ -\frac{1}{|x|} \sum_{t=1}^{|x|} 
    \log P_{{\bfp}^{(\ell)}}\!\left(x_t \;\middle|\; 
    R^{(\ell)}_{\bft}(x_{<t})\right) \right].
\end{align}
For a fixed model parameter $\bft$, we define the optimized probe as $\bfp^\star(\bft)=\arg\min_{\bfp} \mathcal{L}_{p}(\bft, \bfp)$ which extracts the most sensitive information from the probed model. We then penalize the model according to the amount of information recoverable by this optimized probe. This gives the following probing-guided unlearning
objective:
\begin{align} \label{eq:minimax_objective}
    \min_{\bft} \left[
    \mathcal{L}_{f}(\bft) + 
    \alpha \mathcal{L}_{r}(\bft)
    - \frac{\beta}{|L_{p}|} \mathcal{L}_{p}(\bft, \bfp^\star(\bft))\right],
\end{align}
where $\beta$ controls the strength of representation-level suppression. We note that the negative sign for the probe loss indicates that a smaller loss corresponds to greater recoverability of forget content. Therefore, the problem in~\eqref{eq:minimax_objective} encourages the model to increase the optimized probe loss, making sensitive content less recoverable from the probed hidden
representations while still preserving retain-set utility. In our experiments, we exploit the NPO loss~\cite{zhang2024negative} as the forget objective $\mathcal{L}_{f}(\cdot)$ with a cross-entropy as retain loss $\mathcal{L}_{r}(\cdot)$.

\subsection{Experiments}
We evaluate our proposed unlearning method PARS on two datasets, TOFU and MUSE-News.

\noindent\textbf{PARS Training Configuration.} We provide detailed setups in \textbf{Appendix~\ref{sec:train_details}} with models, hyperparameter choices and computational overheads discussions.  

\noindent\textbf{Evaluation Tasks.}
We measure unlearning effectiveness using three 
metrics for each benchmark. For TOFU, we use 
Entailment Score (ES) on $\mathcal{D}_f$, 
Membership Inference Attack (MIA)~\cite{shokri2017membership} exploiting 
Min-K\%++~\cite{shi2023detecting}, and 
$\widehat{\text{leak@}64}$--ES measuring the leakage over $64$ probabilistic generation samples~\cite{reisizadeh2025leak}. For MUSE-News, we report VerbMem and KnowMem on $\mathcal{D}_f$ measuring verbatim and knowledge memorization at the output level, and $\widehat{\text{leak@}64}$--RS. We also evaluate robustness to relearning attacks by measuring ES for TOFU, and VerbMem and KnowMem for MUSE-News, on $\mathcal{D}_f$ after fine-tuning the unlearned model on a small subset of this set~\cite{fan2025towards}. Further, we evaluate unlearning robustness to jailbreak attacks using Greedy Coordinate Gradient (GCG)~\cite{zou2023universal}. The model’s overall utility is evaluated by ES on Retain QA for TOFU and by KnowMem on $\mathcal{D}_r$ for MUSE-News. The details on the evaluation tasks are in  \textbf{Appendix~\ref{sec:eval_details}}. We compare PARS against GradDiff~\cite{liu2022continual}, NPO~\cite{zhang2024negative}, SimNPO~\cite{fan2024simplicity}, SURE~\cite{zhang2024catastrophic}, and SAM~\cite{fan2025towards}. 

\begin{table}[H]
  \centering
  \resizebox{0.8\columnwidth}{!}{
  \begin{tabular}{lccc cc cc}
  \toprule
  \multirow{2}{*}{\textbf{Model}} &
  \multicolumn{3}{c}{\textbf{Unlearning Effectiveness}} &
  \multicolumn{1}{c}{\textbf{Relearning Robustness}} &
  \multicolumn{1}{c}{\textbf{Jailbreak}} &
  \multicolumn{1}{c}{\textbf{Utility}} \\
  \cmidrule(lr){2-4} \cmidrule(lr){5-5} \cmidrule(lr){6-6} \cmidrule(lr){7-7}
  & ES $\downarrow$ & MIA $\downarrow$ & $\widehat{\text{leak@$64$}}$--ES $\downarrow$ &
    ES $\downarrow$ &
    GCG $\downarrow$ & Retain QA $\uparrow$ \\
  \midrule
  Original   & 26.7 & -1.8 & 84.3 & N/A & 0.306 & 30.9 \\
  GradDiff   & 22.4 & -6.3 & 80.5 & 31.9 & 0.260 & 35.0 \\
  NPO        & 18.6 & -3.6 & 75.9 & 20.4 & 0.219 & 29.6 \\
  SimNPO     & 15.6 & -5.4 & 76.4 & 22.4 & 0.237 & 32.2 \\
  RMU        & 17.4 & -3.1 & 96.7 & 26.4 & 0.241 & 35.4 \\
  \rowcolor{blue!20} PARS & \textbf{13.0} & \textbf{-104.6} & \textbf{73.7} & \textbf{20.4} & \textbf{0.198} & 34.7 \\
  \bottomrule
  \end{tabular}}
  \caption{\small Comparison of unlearning performance across baseline methods on the TOFU benchmark. Lower MIA scores indicate diminished memorization and
  superior unlearning efficacy, while lower Leak@64 values signify more thorough knowledge erasure. Robustness against relearning attacks is quantified by
  post-relearning ES, robustness against GCG jailbreak attacks is measured by post-GCG ROUGE-L (lower, more robust), and model utility is assessed via ES on the retain QA set.
  }
  \label{tab:tofu}
  \end{table}
  \vspace{-0.6cm}
  \begin{table}[H]
  \centering
  \resizebox{\columnwidth}{!}{
  \begin{tabular}{lccc ccc c c}
  \toprule
  \multirow{2}{*}{\textbf{Method}} &
  \multicolumn{3}{c}{\textbf{Unlearning Effectiveness}} &
  \multicolumn{3}{c}{\textbf{Relearning Robustness}} &
  \multicolumn{1}{c}{\textbf{Jailbreak}} &
  \multicolumn{1}{c}{\textbf{Utility}} \\
  \cmidrule(lr){2-4} \cmidrule(lr){5-7} \cmidrule(lr){8-8} \cmidrule(lr){9-9}
  & VerbMem $\mathcal{D}_f$ $\downarrow$ & KnowMem $\mathcal{D}_f$ $\downarrow$ & $\widehat{\text{leak@$64$}}$--RS $\downarrow$ &
    VerbMem $\mathcal{D}_f$ $\downarrow$ &
    KnowMem $\mathcal{D}_f$ $\downarrow$ &
    & GCG $\downarrow$ & KnowMem $\mathcal{D}_r$ $\uparrow$ \\
  \midrule
  Original  & 56.3 & 63.7 & 96.4 & N/A & N/A &  & 0.175 & 55.2 \\
  GradDiff  & 22.5 & 53.4 & 93.2   & 73.2 & 57.1 &  & 0.170 & 42.0 \\
  NPO  & 0.3 & 41.3 & 90.9 & 41.5 & 50.1 &  & 0.113 & 41.8 \\
  SimNPO & 13.9 & 47.1 & 92.0 & 78.9 & 52.7 &  & 0.109 & 40.3 \\
  SURE      & 0.0 & 38.7 & 88.7   & 39.5 & 49.5 &  & 0.112 & 40.7 \\
  SAM       & 0.0 & 42.3 & 83.3 & 41.3 & 50.1 &  & 0.096 & 42.6 \\
   \rowcolor{blue!20} PARS & 0.9 & 40.1 & \textbf{80.6} & \textbf{26.8} & \textbf{45.3} &  & \textbf{0.093} & 41.7 \\
  \bottomrule
  \end{tabular}}
  \caption{\small Comparison of unlearning performance across baselines on the MUSE-News set. Lower pre-attack $\widehat{\text{leak@64}}$--RS and pre- and post-attack VerbMem and KnowMem on $\mathcal{D}_f$ indicate more effective unlearning.
  }
  \label{tab:muse}
  \vspace{-0.3cm}
  \end{table}

\noindent\textbf{Results.}
\textbf{Table~\ref{tab:tofu}} reports the performance of PARS and other unlearning baselines on TOFU forget10. PARS outperforms all baselines in terms of the forgetness metric while maintaining comparable performance on the retain task. Moreover, this model achieves a much lower Min-K\%++ score than the evaluated baselines. PARS outperforms RMU on all forget metrics despite RMU being a representation-level approach. Hence, generically perturbing hidden activations, as in RMU, does not necessarily suppress sensitive information encoded in hidden states. Instead, PARS uses an optimized probe to identify information that is decodable from hidden states and trains the model to reduce this probe-based recoverability. We compare PARS with other baselines on MUSE-News in \textbf{Table~\ref{tab:muse}}. While PARS achieves comparable VerbMem and KnowMem prior to relearning attacks, it outperforms \textit{all} baselines on the robustness evaluation metrics, including $\widehat{\text{leak@}64}$--RS, post-relearning VerbMem and KnowMem, and jailbreak robustness. Specifically, relative to SAM, which is \textit{designed} to enhance robustness against relearning attacks, PARS reduces VerbMem by $\textbf{61\%}$ and KnowMem by $\textbf{19.0\%}$ post-relearning. This indicates that suppressing probe-recoverable information from hidden states provides further robustness beyond sharpness-aware optimization.

%% file: sections/conclusion.tex
\vspace{-0.1cm}
\section{Conclusion}
We showed that existing LLM unlearning methods suffer from a fundamental limitation: an illusion of forgetting. Although sensitive outputs are suppressed at the decoder level, the hidden representations retain the underlying knowledge. We formalized this gap through an information-theoretic analysis and then validated it empirically using generative probing across three benchmarks. Finally, we proposed PARS, a probe-adversarial unlearning objective that directly targets representational leakage and outperforms existing baselines across various LLM unlearning tasks, including MIA, Leak@k, relearning, and jailbreak attacks. Our work provides a new lens for understanding information leakage in LLM unlearning and establishes a representation-level foundation for future algorithm design.

%% file: sections/appendix.tex
\section{Related Works}\label{sec:related-work}
\textbf{Adversarial Attacks for Unlearning.} As more LLM unlearning methods are proposed, evaluating their robustness has emerged critical. Various adversarial attacks have been developed to extract undesired information, assessing whether an unlearning approach is truly effective or solely masking the data~\cite{hu2024unlearning}. Current adversarial attacks generally fall into two broad categories:

(1) \textit{Generation-based attacks} exploit the unlearned model's text generation capabilities to recover information. This includes relearning attacks, where lightweight fine-tuning on even a small subset of forget samples can effectively restore unlearned knowledge~\cite{deeb2024unlearning,fan2025towards,lucki2024adversarial}. Another example is jailbreaking attacks, in which adversarial prompts are carefully crafted to bypass safety alignments and recover forgotten information during inference~\cite{zou2023universal,lucki2024adversarial,patil2023can,lynch2024eight}. Recently, metrics such as Leak@k~\cite{reisizadeh2025leak} have been proposed to assess the risk of adversarial, repetitive prompting under probabilistic decoding, aiming to extract worst-case leaked outputs.

(2) \textit{Internal-state-based attacks} target to detect traces of sensitive training data from a model’s internal representations. Instead of relying on generated text, these methods analyze internal behavior, probability distributions, or latent states to determine whether specific training data remains memorized. Techniques such as Min-K\%~\cite{shi2023detecting} and Min-K\%++~\cite{zhang2024min} exploit token probability distributions to detect pre-training data leakage and can be considered a special case of Membership Inference Attacks (MIA)~\cite{shokri2017membership}. Moreover, a linear probe (or auxiliary classifier) is typically a simple linear or MLP model attached to a frozen intermediate representation of a neural network, used to assess what information is encoded in hidden states~\cite{liu2019linguistic,adi2016fine}. In unlearning context, probes are further applied to detect whether sensitive or task-relevant knowledge is still retained in intermediate representations~\cite{li2024wmdp}.

Our framework extends the classification-based probes to a generation-aware probe operating over hidden states and vocabulary distributions. It directly evaluates how much memorized content can be recovered from each layer’s representation. This enables our proposed algorithm, PARS, to quantify not only whether information is linearly separable, but also whether it is generatively decodable, providing a more direct mechanism for identifying and mitigating latent knowledge leakage in LLMs.

\textbf{Adversarial Training Perspectives in Unlearning.} Such training has been widely studied as a mechanism for improving model robustness, though it often incurs substantial computational overhead and optimization challenges~\cite{zhang2019theoretically,madry2017towards}. Recently, machine unlearning approaches have incorporated adversarial paradigms to defend against extraction and relearning attacks. A promising line of work formulates unlearning as a latent-space adversarial min--max problem, in which perturbations are introduced to hidden states and optimized to recover forgotten knowledge, while model parameters are updated to resist such worst-case activations~\cite{sheshadri2024latent,yuan2025towards,li2026textbfagtaorobuststabilizedllm}.

Another direction of work~\cite{fan2025towards} models relearning attacks through a parameter-space min--max formulation, where worst-case weight perturbations are constructed to reverse unlearning, and robustness is enforced via sharpness-aware minimization to promote a smooth loss landscape. Despite these advances, such methods primarily assess robustness through output-level recovery, without explicitly probing or constraining residual sensitive information within intermediate representations.

To address this issue, the work~\cite{yan2025dual} proposes a probe-based adversarial framework (PRISM), which trains a latent probe to distinguish harmful and safe representations under adversarial perturbations, and subsequently guides unlearning by pushing harmful representations toward a probe-defined safe region. However, this approach relies on the probe as a surrogate objective, enforcing decision-boundary alignment rather than explicitly identifying and removing sensitive information embedded in representations.

In contrast, our paradigm directly operates on intermediate representations by probing layer-wise hidden states with generative supervision, enabling it to capture and expose residual sensitive information that may not be reflected in classification boundaries. By aligning these representations with target generations rather than binary decisions, PARS can more precisely identify and erase latent leakage beyond probe-level classification.

\section{Experimental Details}\label{sec:exp-details}
\subsection{Evaluation Tasks}\label{sec:eval_details}

\textbf{MIA Min-K\%++} quantifies memorization using normalized token log-likelihoods. For each token $x_t$, the score is
\begin{align*}
    \text{Min-K\%++}_{\text{token}}(x_t)
= \frac{\log p(x_t \mid x_{<t}) - \mu_{x_{<t}}}{\sigma_{x_{<t}}},
\end{align*}
where $\mu_{x_{<t}}$ and $\sigma_{x_{<t}}$ are the mean and standard deviation of log-probabilities over the vocabulary conditioned on $x_{<t}$. The sequence-level score averages the lowest $k\%$ token scores:
\begin{align*}
    \text{Min-K\%++}(x) = \frac{1}{|{\min\text{-}k\%}|} \sum_{x_t \in \min\text{-}k\%} \text{Min-K\%++}_{\text{token}}(x_t).
\end{align*}
A higher score indicates stronger memorization, while a lower score indicates better unlearning on the forget set. In our experiment, we use $k=40\%$ (i.e., the lowest 40\% of tokens).

\textbf{Leak@k} evaluates information leakage under adversarial multiple prompting under probabilistic decoding. In our experiment, Leak@k generates $n = 200$ independent samples to measure the reappearance of forgotten knowledge within $k = 64$ generations. Leak@k employs probabilistic decoding, for which we set the temperature to $1.0$ and top-$p$ to $1$. For the dataset-specific evaluation metrics, we utilize RS-Recall for the MUSE dataset and ES for the TOFU dataset.

\textbf{Relearning Attacks} evaluate the robustness of the unlearning process by fine-tuning the unlearned model on the forget set for a short duration. This tests the model's true forget quality by measuring how easily the supposedly erased information can be recovered after this brief relearning phase. The specific hyperparameters utilized for the relearning attacks across the TOFU and MUSE datasets are detailed in \textbf{Table~\ref{tab:relearning_hyperparams}.}

\textbf{Jailbreak Attacks} evaluate whether adversarial prompts can extract the sensitive content from the unlearned model. We use Greedy Coordinate Gradient (GCG)~\cite{zou2023universal} to optimize a shared adversarial suffix $s$ on a subset $\mathcal{D}_a \subset \mathcal{D}_f$, given as
\begin{align*}
s^{*} = \arg\min_s -\frac{1}{|\mathcal{D}_a|}
      \sum_{(x,y)\in\mathcal{D}_a}
      \frac{1}{L_y}\sum_{t=1}^{L_y}
      \log p_{\bft_u}(y_t \mid x \oplus s, y_{<t}),
\end{align*}
where $\bft_u$ denotes the unlearned model, $x$ is the input prompt, $y$ is the undesired target, $\oplus$ denotes concatenation, and $L_y=\min(16,|y|)$ restricts optimization to the target prefix. GCG iteratively proposes token
substitutions using gradients and selects the suffix with the lowest average target loss. We then append the optimized suffix to held-out forget prompts and evaluate extraction under greedy decoding; greater extraction indicates weaker robustness to jailbreak attacks.

\begin{table}[h]
    \centering
    \begin{tabular}{lccc}
        \toprule
        \textbf{Dataset} & \textbf{Learning Rate} & \textbf{Batch Size} & \textbf{Steps} \\
        \midrule
        TOFU & $2\times 10^{-5}$ & $8$ & $75$ \\
        MUSE & $2\times 10^{-5}$ & $4$ & $75$ \\
        \bottomrule
    \end{tabular}
    \caption{Hyperparameters for Relearning Attacks.}
    \label{tab:relearning_hyperparams}
\end{table}
\subsection{PARS Training Setup.}\label{sec:train_details}
We used NVIDIA A100 to perform full parameter unlearning. Specifically, we used $8$ $\times$ A100 to run unlearning for MUSE-News benchmark, $2$ $\times$ A100 to run unlearning for TOFU benchmark.

\noindent\textbf{PARS Hyperparameters.} On TOFU and MUSE benchmarks, PARS is implemented on the NPO loss. For TOFU, it uses $6$ epochs, learning rate of $10^{-5}$, batch size of $32$, $\alpha=6$, and $\beta=0.5$. For MUSE, we exploit $6$ epochs, learning rate of $10^{-5}$, batch size of $2$, $\alpha=4$, and $\beta=0.75$.

\noindent\textbf{Computational Overheads.}
Our PARS algorithm introduces an additional probing loss to facilitate the erasure of internal knowledge within intermediate layers. While this approach yields significantly more robust performance across various attacks, it inherently introduces a computational tradeoff. Specifically, the auxiliary loss requires additional GPU memory to store intermediate representations and gradients. For instance, utilizing four intermediate layers incurs an estimated additional GPU memory overhead of $15\%$ for MUSE-News and $50\%$ for TOFU. Furthermore, the training time increases compared to standard NPO training: from 1 hour to 3 hours on TOFU, and from $2$ hours to $4$ hours on MUSE-News, see in \textbf{Table~\ref{tab:computational_overheads}}.
\begin{table}[H]
    \centering
    \resizebox{\linewidth}{!}{
    \begin{tabular}{lccccc}
        \toprule
        \textbf{Dataset} & \textbf{Model} & \textbf{Target Layers ($L_p$)} & \textbf{Baseline Time} & \textbf{PARS Time} & \textbf{Extra GPU Mem.} \\
        \midrule
        TOFU & LLaMA-3.2-1B-Instruct & $\{8, 10, 12, 14\}$ & $2.0$ hr & $4.0$ hrs & $+50\%$ \\
        MUSE-News & LLaMA2-7B & $\{24, 26, 28, 30\}$ & $1.0$ hr & $3.0$ hrs & $+15\%$ \\
        \bottomrule
    \end{tabular}}
    \caption{Summary of dataset-specific configurations and computational overheads for the PARS algorithm compared with baseline time refers to standard NPO training.}    \label{tab:computational_overheads}
\end{table}

\section{Proof of Theorem~\ref{thm:linear_non}}\label{sec:proof-theorem}
We first present an auxiliary lemma.

\begin{lemma}
\label{lem:softmax_KL_quadratic}
Let $q_x := \mathrm{Categorical}(\mathrm{softmax}(x))$ for $x \in \mathbb{R}^n$.
Then, for any $z,\eta\in \mathbb{R}^n$,
\begin{align}\label{eq:softmax_KL_quad}
\mathrm{KL}(q_{z}\|q_{z+\eta})\ \le\ \frac{1}{2}\,\|\eta\|_\infty^2.
\end{align}
\end{lemma}
The proof of Lemma~\ref{lem:softmax_KL_quadratic} is provided in \textbf{Appendix~\ref{sec:proof-lem}}. Now, we are ready for the proof of Theorem~\ref{thm:linear_non}. We prove (a) and (b) separately.

\noindent\textbf{(a)} Let $T:=v^\top(R-r_0)$.
Then, using $R=r_0+(\mu_S+\zeta)v+\xi$, we have
\begin{align*}
T& = v^\top\big((\mu_S+\zeta)v+\xi\big)\nonumber\\
& = v^\top v (\mu_S+\zeta) + v^\top \xi = \mu_S+\zeta
\end{align*}
where the last step follows from $\|v\|_2=1$ and $v^\top \xi = 0$. Since $T$ is a measurable function of $R$, the data processing inequality gives
\begin{align}\label{eq:ISRX_ge_IST}
    I(S;R)\ &\ge\ I(S;T) = h(T)-h(T\mid S).
\end{align}
For the second term in~\eqref{eq:ISRX_ge_IST}, since conditioning on $S$ shifts $T$ by the constant $\mu_S$, we have
\begin{align}\label{eq:h_T_S}
   h(T\mid S)=h(\zeta)=\frac12\log(2\pi e\,\tau^2).
\end{align}
Since $\mu_S$ is a measurable function of $S$ and $\zeta\perp S$, we have $\mu_S\perp \zeta$. Hence, to lower bound the first term in~\eqref{eq:ISRX_ge_IST}, applying the entropy power inequality, we get
\begin{align}\label{eq:N_T}
\mathsf N(T) & = \mathsf N(\mu_S+\zeta) \geq \mathsf N(\mu_S)+\mathsf N(\zeta)
=\mathsf N(\mu_S)+\tau^2,
\end{align}
where the last step follows from $\zeta\sim \mathcal N(0,\tau^2)$. From~\eqref{eq:N_T}, we can write
\begin{align}\label{eq:h_T_X}
    h(T)\ \ge\ \frac12\log\!\big(2\pi e(\mathsf N(\mu_S)+\tau^2)\big).
\end{align}
Substituting~\eqref{eq:h_T_S} and~\eqref{eq:h_T_X} into~\eqref{eq:ISRX_ge_IST}, we obtain
\begin{align*}
I(S;R)\ \ge\ \frac12\log\!\left(1+\frac{\mathsf N(\mu_S)}{\tau^2}\right).
\end{align*}

\noindent\textbf{(b)} Fix $c\in\mathcal C$. For each $s\in\mathrm{supp}(S\mid C=c)$, we define $p_s := P(Z\in\cdot \mid C=c,S=s)$ and $\bar p := P(Z\in\cdot\mid C=c)=\mathbb E_{S\mid C=c}[p_S]$. Using the identity for conditional mutual information, we have
\begin{align}\label{eq:I_as_KL_b}
I(S;Z\mid C=c) = \mathbb E_{S\mid C=c}\!\left[\mathrm{KL}(p_S\|\bar p)\right].
\end{align}
Since KL divergence is convex in its second argument, for each fixed $s$, we get
\begin{align*}
\mathrm{KL}\!\left(p_s\Big\|\mathbb E_{S'\mid C=c}[p_{S'}]\right) \le
\mathbb E_{S'\mid C=c}\!\left[\mathrm{KL}(p_s\|p_{S'})\right],
\end{align*}
where $S'$ is an independent copy of $S$ under $P_{S\mid C=c}$.
Averaging over $S\mid C=c$, we obtain 
\begin{align}\label{eq:I_pairwise_b}
I(S;Z\mid C=c) \le \mathbb E_{S,S'\mid C=c}\!\left[\mathrm{KL}(p_S\|p_{S'})\right].
\end{align}
Let $\Theta := (\zeta,\xi)$ denote the noise variable, independent of $(C,S)$. 
For each $s$, we define the $\Theta$-conditional distribution as $p_{s,\theta} := P(Z\in\cdot \mid C=c, S=s, \Theta=\theta)$. Under the model, $p_{s,\theta}=\mathrm{Categorical}(\mathrm{softmax}(V_{s,\theta}))$ with logits $V_{s,\theta}
= W\big(r_0 + (\mu_s+\zeta)v+\xi\big)+b$ and $\mu_s := \mathbb E[\alpha\mid S=s]$. Moreover, we can write $p_s = \mathbb E_{\Theta}[p_{s,\Theta}]$. From the joint convexity of KL divergence, we can conclude
\begin{align}\label{eq:KL_joint_convex}
\mathrm{KL}(p_s\|p_{s'}) = \mathrm{KL}\!\left(\mathbb E_{\Theta}[p_{s,\Theta}]  \,\Big\|\,  \mathbb E_{\Theta}[p_{s',\Theta}] \right) \le \mathbb E_{\Theta}\!\left[\mathrm{KL}(p_{s,\Theta}\|p_{s',\Theta})\right].
\end{align}
For fixed $\theta=(\zeta,\xi)$, the logit difference satisfies $V_{s,\theta}-V_{s',\theta} = (\mu_s-\mu_{s'})\,Wv$. Since $v$ is $\varepsilon$-insensitive, there exist 
$\lambda^\star\in\mathbb R$ and $u\in\mathbb R^n$ with  $\|u\|_\infty\le \varepsilon$ such that $Wv=\lambda^\star\mathbf 1+u$. Because $\mathrm{softmax}(z+c\mathbf 1)=\mathrm{softmax}(z)$ for any $c\in\mathbb R$,  the additive shift $\lambda^\star\mathbf 1$ does not affect the induced categorical distribution. Hence we may equivalently compare logits differing by $\eta := (\mu_s-\mu_{s'})\,u$ with $\|\eta\|_\infty \le \varepsilon\,|\mu_s-\mu_{s'}|$. Applying Lemma~\ref{lem:softmax_KL_quadratic}, we obtain
\begin{align}\label{eq:I_3}
    \mathrm{KL}(p_{s,\theta}\|p_{s',\theta}) \le  \frac12\,\|\eta\|_\infty^2 \le  \frac12\,\varepsilon^2(\mu_s-\mu_{s'})^2.
\end{align}
Taking expectation~\eqref{eq:I_3} over $\Theta$ and using \eqref{eq:KL_joint_convex}, we get
\begin{align}\label{eq:I_4}
    \mathrm{KL}(p_s\|p_{s'}) \le  \frac12\,\varepsilon^2(\mu_s-\mu_{s'})^2.
\end{align}
Finally, taking expectation in~\eqref{eq:I_4} over i.i.d.\ $(S,S')$
drawn from $P_{S\mid C=c}$ and using~\eqref{eq:I_pairwise_b}, we get
\begin{align}\label{eq:I_5}
I(S;Z\mid C=c)
&\le \frac12\,\varepsilon^2\,
\mathbb E\big[(\mu_S-\mu_{S'})^2 \mid C=c\big] = \varepsilon^2\,\Var(\mu_S\mid C=c).
\end{align}
Averaging~\eqref{eq:I_5} over $C$, we yield
\begin{align*}
I(S;Z\mid C)
\le \varepsilon^2\,\mathbb E[\Var(\mu_S\mid C)]
\le \varepsilon^2\,\Var(\mu_S),
\end{align*}
where the last step follows from the law of total variance. This completes the proof of (b).

\section{Proof of Lemma~\ref{lem:exact}}\label{sec:proof-exact}
From the definition of $r_0$, we have $R=r_0+\beta$ where $\beta:=R-\mathbb E[R]$. Fix a unit vector $v\in\mathbb R^d$ and let $P:=vv^\top$ and $P_\perp:=I-P$. Hence, we can write 
\begin{align}\label{eq:beta}
    \beta=P\beta+P_\perp\beta=(v^\top\beta)v+P_\perp\beta.
\end{align}
We define $\alpha:=v^\top(R-r_0)=v^\top\beta$ and $\xi:=P_\perp(R-r_0)=P_\perp\beta$. Using~\eqref{eq:beta}, we get
\begin{align}\label{eq:R_0}
    R &= r_0 + (v^\top\beta)v+P_\perp\beta =r_0+\alpha v+\xi.
\end{align}
Now, we decompose $\alpha$ as $\alpha=\mu_S+\zeta$ where $\mu_S:=\mathbb E[\alpha\mid S]$ and $\zeta:=\alpha-\mathbb E[\alpha\mid S]$. This together with~\eqref{eq:R_0} leads us to
\begin{align*}
    R=r_0+(\mu_S+\zeta)v+\xi,
\end{align*}
with $\mathbb E[\zeta\mid S]=0$. Moreover, using the fact $\|v\|_2=1$, we have
\begin{align*}
    v^\top\xi = v^\top(I-vv^\top)(R-r_0) = (v^\top-v^\top vv^\top)(R-r_0)=0.
\end{align*}

\section{Proof of Lemma~\ref{lem:softmax_KL_quadratic}}\label{sec:proof-lem}
We define the log-partition function $g$ as
\begin{align*}
    g(z):=\log\sum_{i=1}^n e^{z_i}, \qquad z\in\mathbb R^n.
\end{align*}
Then, $g$ is $C^\infty$ and convex, with
\begin{align*}
    \nabla g(z)=q,\qquad \nabla^2 g(z)=\mathrm{Diag}(q)-qq^\top,
\end{align*}
where $q=\mathrm{softmax}(z)$. In particular, for any $\eta\in\mathbb R^n$, we have
\begin{align*}
    \eta^\top \nabla^2 g(z)\,\eta = \eta^\top(\mathrm{Diag}(q)-qq^\top)\eta = \sum_{i=1}^n q_i \eta_i^2 \!-\!\Big(\sum_{i=1}^n q_i\eta_i\Big)^2 = \mathrm{Var}_{i\sim q}(\eta_i).
\end{align*}
The softmax family $\{q_z\}_{z\in\mathbb R^n}$ is an exponential family with natural parameter $z$ and log-partition function $g$; hence its Kullback--Leibler divergence satisfies the standard identity
\begin{align}\label{eq:lm-0}
    \mathrm{KL}(q_z\|q_{z+\eta}) = g(z+\eta)-g(z)-\langle \nabla g(z),\eta\rangle = D_g(z+\eta,z),
\end{align}
i.e., it equals the Bregman divergence of $g$ at $(z+\eta,z)$. By Taylor's theorem in integral form, we get
\begin{align}\label{eq:lm-1}
    D_g(z+\eta,z) = \int_0^1 (1-t)\, \eta^\top \nabla^2 g(z+t\eta)\,\eta\,dt.
\end{align}
For any distribution, $\mathrm{Var}(X)\le (\mathrm{range}(X))^2/4$.
Therefore, for each $t\in[0,1]$, we arrive at
\begin{align}\label{eq:lm-2}
    \eta^\top \nabla^2 g(z+t\eta)\,\eta \!=\! \mathrm{Var}_{i\sim \mathrm{softmax}(z+t\eta)}(\eta_i) \leq \frac{(\max_i \eta_i\!-\!\min_i \eta_i)^2}{4} \!\leq\! \|\eta\|_\infty^2,
\end{align}
where the last inequality follows from $\max_i\eta_i-\min_i\eta_i\le 2\|\eta\|_\infty$. Substituting~\eqref{eq:lm-1} into~\eqref{eq:lm-0}, we arrive at
\begin{align*}
    \mathrm{KL}(q_z\|q_{z+\eta}) & = \int_0^1 (1-t)\, \eta^\top \nabla^2 g(z+t\eta)\,\eta\,dt \stackrel{\rm{(a)}}{\leq} \int_0^1 (1-t)\, \|\eta\|_\infty^2\, dt  = \frac{1}{2}\,\|\eta\|_\infty^2,
\end{align*}
where $\rm{(a)}$ holds due to~\eqref{eq:lm-2}. This completes the proof of the lemma.